\documentclass[letterpaper]{article}
\usepackage{arxiv_preprint}
\usepackage[hyphens]{url}
\usepackage{graphicx}
\usepackage{natbib}
\usepackage{caption}
\usepackage{algorithm}
\usepackage{algorithmic}
\usepackage{amsmath}
\usepackage{amssymb}
\usepackage{booktabs}
\usepackage[table]{xcolor}
\usepackage{multirow}
\usepackage{makecell}
\usepackage{bm}

\copyrighttext{Preprint}

\title{Energy-Structured Latent World Models with Neural Time Fields for Physically Constistent Open-World Motion Planning}

\author{
    Yapeng Liu\textsuperscript{\rm 1,\rm 2,\rm 3},
    Yuanzhao Zhai\textsuperscript{\rm 1,\rm 2},
    Bo Ding\textsuperscript{\rm 1},
    Huaimin Wang\textsuperscript{\rm 1,\rm 2},
    Lin Wang\textsuperscript{\rm 3}
}
\affiliations{
    \textsuperscript{\rm 1}PDL Lab, College of Computer Science and Technology, National University of Defense Technology\\
    \textsuperscript{\rm 2}State Key Laboratory of Complex \& Critical Software Environment, Changsha 410073, Hunan, China\\
    \textsuperscript{\rm 3}EmPACT Lab, Nanyang Technological University, Singapore
}

\newcommand{\Ham}{\mathcal{E}_\theta}

\newcommand{\sgeo}{s_{\mathrm{geo},t}}
\newcommand{\seff}{s_{\mathrm{eff},t}}
\newcommand{\system}{PC-NTF}

\begin{document}

\maketitle

\begin{abstract}
Physically consistent motion planning remains a fundamental challenge in embodied AI, as generated trajectories must strictly conform to real-world execution dynamics. While latent world models offer a promising approach by predicting these dynamics, existing methods learn unconstrained future representations where absorbed physics remains implicit. Therefore, they fail to form reusable physical knowledge, which compromises reliability in unpredictable open-world navigation. To address this, we propose a novel Energy-Structured Latent World Model (\textbf{ELWM}). Our \textbf{key idea} is to structure the ELWM latent state to explicitly carry energy and momentum, ensuring strictly causal transitions via dissipation and control ports. Trained on multimodal RGB-D and inertial interaction histories, our model guarantees physically consistent predictions. We further implement this for motion planning by constructing Physics-Conditioned Neural Time Fields (\textbf{\system{}}), a \textbf{key technical cornerstone} that integrates ELWM into an arrival time field via the Eikonal equation to yield a physically-informed navigation policy. Across held-out scenes, our evaluation reveals significant improvements. Compared to generic latent models, \system{} reduces 0.8-s motion-prediction NRMSE from 0.36 to 0.29. Against Active Neural Time Fields, it improves navigation success from 81.3\% to \textbf{89.7\%} and SPL from 0.64 to \textbf{0.73}, while cutting the physical collision rate from 12.1\% to \textbf{5.8\%} and the Eikonal residual from 0.083 to \textbf{0.031}. Beyond these targeted gains, our results demonstrate that embedding explicit physical structures into latent spaces intrinsically bridges the gap between predictive world models and safe, dynamically feasible motion planning. 
\end{abstract}

\section{Introduction}

Physically consistent motion planning is a fundamental challenge in embodied AI, the trajectory is useful only when the robot can execute it under its current dynamics \cite{hsu2002randomized,liu2025aligning}. The collision-free geometric path may become unsafe or inefficient when dissipation payload, actuation limits, or the robot's dynamic state changes, and these factors often have weak geometric signatures and become observable through interaction \cite{pham2013kinodynamic,bohg2017interactive,liu2026siam}. Physical consistent motion planning in open-world environment therefore requires a predictive model of how the robot--environment system responds to execution \cite{ha2018recurrent,kumar2021rma,huang2026h}.

\begin{figure}[t!]
    \centering
    \includegraphics[width=1\columnwidth]{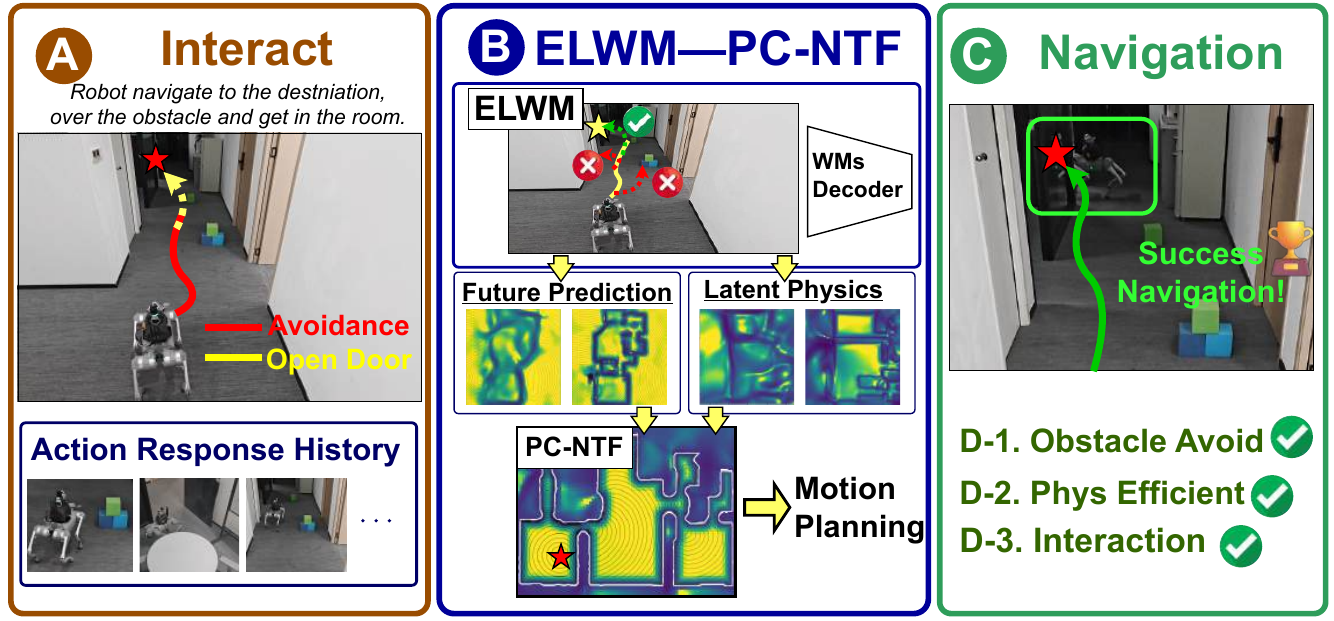}
    \caption{\textbf{ELWM--PC-NTF} bridges action-conditioned physical prediction and motion planning. \textbf{(A)} Robot interactions need action--response capability, \textbf{(B)} ELWM predicts future motion and latent physical dynamics, and PC-NTF combines them with neural time fields. \textbf{(C)} System closed-loop execution produces obstacle-aware, physics-informed navigation towards the goal.}
    \label{fig:intro-teaser}
    \vspace{-0.4cm}
\end{figure}

Latent world models provide a promising basis for this predictive component \cite{lecun2022path,bar2025navigation}. By encoding multi-modal observation and interaction histories into latent states, world models support action-conditioned rollouts without reconstructing every future observation \cite{hafner2019learning,hafner2019dream,hafner2023mastering}. Recent models extend this principle to motion planning and navigation \cite{zhou2025dino,chen2025vl}.
However, latent prediction does not imply reusable physical knowledge or formal physical consistency \cite{lutter2019deep,shang2026roboscape}. Existing methods mostly optimize unconstrained future representations, leaving dissipation, interaction response, and other physical regularities entangled in task-, env- and embodiment-specific transition weights \cite{ye2026world,gong2026csympnet,chen2026lawam}. They expose neither a causal physical structure nor a predicted quantity that motion planner can generally consume. This limitation is especially consequential in open-world navigation, where a planner must adapt its decision to physical interactions that are not determined by scene geometry alone.

To make physical dynamics reusable, \emph{we study the latent transition of how to learn transferable physical regularities and effectively implement them in motion planning}. Consequently, we propose the \textbf{Energy-Structured Latent World Model (ELWM)}, which encodes multimodal RGB-D and inertial interaction histories, together with executed trajectories, into the general physical regularities and dynamic latent phase state \cite{desai2021port}. We formalize the latent stored-energy function and structure the explict port-Hamiltonian (pH) transition for navigation dissipation and control ports \cite{zhong2019symplectic}. This construction imposes an action-conditioned, energy-balance-compatible transition structure. Given a candidate command sequence, ELWM rolls this structured state forward and decodes planner-observable consequences, including the predicted pose, velocity, interaction, and traversal progress. The learned structure can therefore be evaluated through both latent future prediction and port-Hamiltonian motion planning dynamics.

ELWM exposes explicit physical interaction consequences, while these predictions do not specify how the robot should progress toward a navigation goal by themselves. The motion planner requires a global value representation which connects locally achievable outcomes to global navigation cost \cite{maes2026leworldmodel}. Neural time fields provide this complementary substrate by interpreting time-to-go as a continuous arrival-time function constrained by the Eikonal equation, and Active Neural Time Fields (ANTF) update the field from the observed geometry of the scene incrementally, the resulting arrival time and its goal-directed gradient efficiently organize global geometric guidance \cite{ni2022ntfields,liu2025physics}. However, ANTF derives this guidance primarily from static geometry map, it induces the same preference even when ELWM predicts different traversal responses under changes in payload or actuation. The missing link is therefore a motion planning system that utilizes ELWM observable, candidate-dependent predictions to condition arrival-time decisions, beyond the mappings from states, poses, or dissipation to an Eikonal propagation speed.

We close this interface gap through \textbf{Physics-Conditioned Neural Time Fields (\system{})} system. The overall motivation are summarized in Figure~\ref{fig:intro-teaser}, ELWM and PC-NTF bridges the physical prediction and motion planning. \system{} decomposes motion planning responsibility between ELWM's local interacting execution prediction and ANTF's global geometric time-to-go envelope. ANTF first supplies the nominal descent direction and a shared set of feasible action candidates. Starting from the same interaction history, the trained ELWM predicts how each candidate will be executed over a short horizon. PC-NTF then queries the neural time field at each predicted terminal position and then fixes the physics conditioned traversal delay derived from the predicted interaction progress from a physics latent decoder. The resulting model-based time-to-go \emph{backup} treats the world-model consequence as an action-conditioned physical stage cost and the neural time field as a geometric terminal value. It thereby connects ELWM to arrival-time planning without interpreting each latent mechanics as the Eikonal equation participant. Because the correction is candidate dependent online, PC-NTF can change the locally executed decision while preserving the global geometric structure of ANTF, without requiring ELWM to predict paths to the goal or the time field to be retrained for each physical interaction condition.

We evaluate this framework along two complementary axes that mirror its two technical components. At the world-model level, ELWM is compared with a capacity-matched generic latent world model in MPC controller using observable multi-step motion NRMSE, action interventions, and Hamiltonian dynamics diagnostics on held-out physical regimes and scenes. At the planning level, PC-NTF is compared with geometry ANTF  and we report navigation success, SPL, physical collision rate, arrival-time accuracy, Eikonal residual, and planning latency. This separation allows the contributions analysis of generic prediction, energy-structured dynamics, and neural time-field guidance to be measured independently.
Experiments show that ELWM obtains motion NRMSEs of $0.705$ and $0.697$ on the held-out regime and scene, respectively, while maintaining a discrete energy-balance residual of $7.12\times10^{-7}$; action shuffling increases latent prediction error by approximately $10\%$.
These results support action-conditioned prediction and consistency of the physical transition, whereas navigation positive improvement from success rate, latent error and Eikonal residual also indicates \system{} motion planning efficiency.

Our contributions follows:
\begin{itemize}
    \item \textbf{ELWM Construction.} We introduce \textbf{ELWM}, an energy-structured latent world model that organizes action-conditioned dynamics through a latent phase state with explicit stored-energy, dissipation, and control structure.
    \item \textbf{PC-NTF System Implementation.} We propose \textbf{PC-NTF}, the motion planning system that utilize ELWM predicted execution consequences with NTF geometric time-to-go through specific terminal states and physically dimensioned traversal delays.
    \item \textbf{Efficient motion planning results.} We evaluate ELWM and PC-NTF through the physical consistent prediction and motion planning in open-world environments. Experiment shows that we reduce 0.8-s motion-prediction NRMSE from 0.36 to \textbf{0.29}, improve navigation success from 81.3\% to \textbf{89.7\%} and SPL from 0.64 to \textbf{0.73}.
\end{itemize}

\section{Related Work}
\noindent \textbf{Physically Consistent Latent World Models.}
Latent world models compress observation action-conditioned histories and future dynamics into meaningful representations, which can support planning and policy learning \cite{ha2018recurrent,hafner2019learning,hafner2019dream,hafner2023mastering}. Recent models extend this principle to pretrained visual features, navigation, and robot policies \cite{zhou2025dino,rao2026skyjepa,chen2026lawam}. These results establish the decision relevance of latent rollouts, but predictive sufficiency does not make robot--environment interaction mechanics reasonable: effects such as payload and dissipation may remain entangled in unconstrained latent transition \cite{nie2026phys,scholkopf2021toward}.
Physical consistent world models were proposed to address this limitation, including physics-informed world models and mechanics-structured networks. The former incorporate differentiable simulators or physical supervision into predictive learning \cite{li2025pin,shang2026roboscape}, whereas the latter constrain the learned transition itself \cite{greydanus2019hamiltonian,zhong2019symplectic}. More recent approaches bring these principles closer to world models construction, AC-HGN embeds actions as external forces in an abstract Hamiltonian phase space \cite{troch2025action}; Phys-JEPA structures physical and residual components of latent prediction \cite{nie2026phys}; and PH-Dreamer combines action-controlled energy flow and dissipation into recurrent world-model dynamics \cite{luan2026ph}.
These advances establish that physical consistent structure can improve the latent prediction and control, while they mainly evaluate with in the learned model or policy. ELWM \textbf{targets on the motion planning}, \emph{extracting the navigation regularities and organize action-conditioned prediction through dissipation}. In this way, ELWM provides the interaction dynamics and shows the efficiency on jointly motion planning.

\noindent \textbf{Neural Time Fields for Motion Planning.}
Efficient motion planning can be characterized by Hamilton--Jacobi and Eikonal equations, with classical solutions obtained by dynamic-programming and Fast Marching methods \cite{lions2006optimal,sethian1996fast}.
For motion planning, neural fields recently have emerged as suitable environment representations due to their compactness and continuity property \cite{park2019deepsdf,you2023generative}, and neural PDE solvers approximate Eikonal solutions continuously through data and equation residuals \cite{raissi2018physics,smith2020eikonet,bin2021pinneik}. NTFields applies this formulation to robot configuration spaces without expert paths \cite{ni2022ntfields}; Active NTFields further learns arrival-time maps online in open-world environments \cite{liu2025physics,zhong2025multi}; and Eikonal caging extends the formulation to contact-rich robot arm manipulation \cite{zhang2026physics}. These methods define physical consistency primarily through the Eikonal PDE and geometry-derived propagation speed, but they did not take the dynamic online interaction consequences in consideration. PC-NTF lifts the arrival-time Eikonal formulation by using \textbf{Practical physical traversal costs} through its decoder and the physics condition, to form the \textbf{online interactive motion planning framework}.

\noindent \textbf{Open-World Motion Planning Under Changing Dynamics.}
Open-world motion planning requires navigation without fully known maps and fixed execution conditions \cite{du2011robot}.
Classical sampling-based planners search for collision-free paths in a given configuration-space representation \cite{karaman2011sampling}, while learning-based planners accelerate this search using learned trajectory proposals and classical feasibility checks \cite{qureshi2020motion}. These methods represent feasibility through geometric collision constraints. Further, experience-driven navigation complements geometric planning by learning local reachability or traversability from robot interaction. BADGR learns navigational affordances from self-supervised experience \cite{kahn2021badgr} and WayFAST uses online traction estimates to supervise terrain-traversability prediction \cite{gasparino2022wayfast}.
However, their planner-facing quantities are typically expressed as collision. These evaluation do not explictly predict how each candidate command sequence will execute under the robot--environment setting. PC-NTF evaluates feasible motion candidates \textbf{under the robot's current interaction response} along with ELWM, allowing open-world decisions to change with payload, dissipation, actuation limits, and dynamic state.

\section{Method}

\begin{figure*}[t!]
    \centering
    \includegraphics[width=0.9\textwidth]{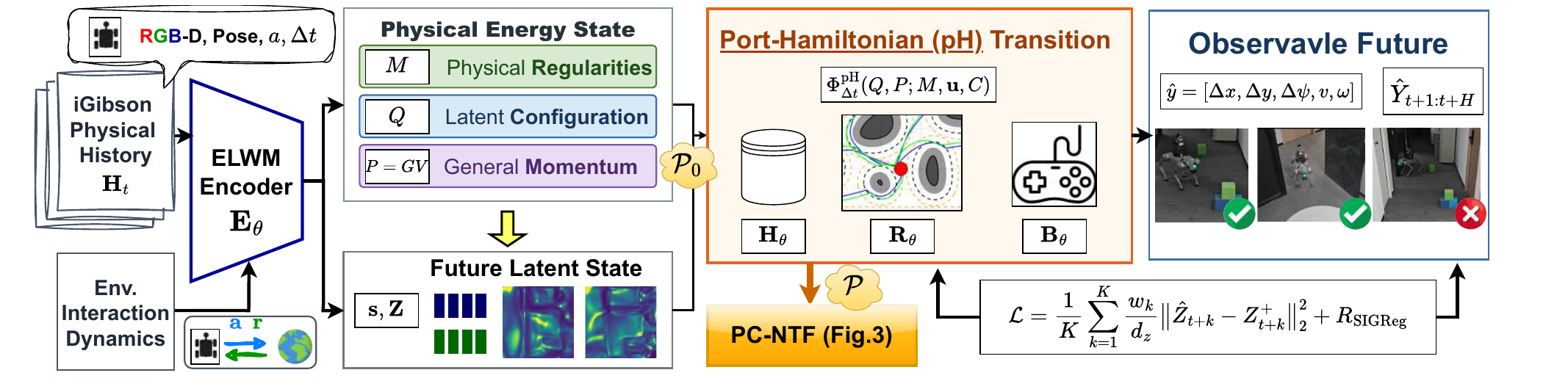}
    \caption{Overview of ELWM. Physical history and interaction dynamics are encoded into physical and latent state; a controlled, dissipative port-Hamiltonian transition formalizes the prediction latent prior as motion futures $\mathcal{P}$ for PC-NTF planning (in Fig.3). ELWM Training by the state future consistency.}
    \label{fig:pipeline}
    \vspace{-0.4cm}
\end{figure*}

\subsection{Preliminary}

\noindent \textbf{Problem Setting.} 
We consider a differential-drive robot that navigates in a planar workspace
$\Omega\subset\mathbb{R}^{2}$ toward the goal $\mathbf{g}$ in Eikonal form, its state can be expressed as
\begin{equation}
\mathbf{r}_t
=
\left(\mathbf{x}_t,\psi_t,\boldsymbol{\xi}_t\right),
\qquad
\boldsymbol{\xi}_t=[v_t,\omega_t],
\end{equation}
and the executed control is
$\bm{u}_t=[v_t^{\mathrm{cmd}},\omega_t^{\mathrm{cmd}}]$.
An upstream navigation task module with VLM foundation model supplies the goal $\mathbf{x}_g$ from dataset, $\mu_r$ denotes robot meta-information (mass/payload, velocity and acceleration limits). Actions $a_t = (\nu_t, \Delta t)$ are the velocity commands received by the base, and $H_t^{\mathrm{phy}} = \{o_\tau^{\mathrm{phy}}, \xi_\tau, a_{\tau-1}, \mu_r\}_{\tau=t-L}^{t}$ collects the recent proprioceptive and navigation interaction history.

\noindent \textbf{Hypothesis.} The physical quantity delivered to the planner is \emph{isotropic traversal-speed field}: within one replanning window, the maximum speed the robot can stably sustain at each location is represented by scalar. Orientation-dependent friction, turning-radius limits, and acceleration state are absorbed into this scalar and enforced at execution time by an explicit constraint layer. For detailed neural time field calculation, we represent them inside through an anisotropic and state-space Hamilton--Jacobi formulation, which we analyze in PC-NTF realization \cite{bardi1997optimal}.

\noindent \textbf{System overview.}
\system{} couples a physical prediction world model with NTF planner through motion planning interface, the inference loop is summarized as
\begin{equation}
\begin{aligned}
\mathbf{H}_t &\xrightarrow{\;\text{ELWM}\;} \mathbf{s}_t,
\nonumber\\
\left(T_{\mathrm{A}},\mathbf{r}_t,\mathbf{g}\right) &\xrightarrow{\;\text{NTF}\;} \left\{\mathbf{U}^{(i)}_t\right\}_{i=1}^{N},
\nonumber\\
\left( \mathbf{s}_t, \left\{\mathbf{U}^{(i)}_t\right\}_{i=1}^{N}, T_{\mathrm{A}} \right) &\xrightarrow{\;\text{PC-NTF}\;} \bm{u}^{\mathrm{exec}}_t \longrightarrow \mathbf{H}_{t+1}.
\label{eq:system-overview}
\end{aligned}
\end{equation}
ELWM summarizes the interaction history as $\mathbf{s}_t$; NTF uses the current state, goal, and arrival-time field $T_{\mathrm{A}}$ to generate shared candidates; and PC-NTF predicts their physical consequences, reranks them with the time field, and executes the selected feasible action. The observed transition then updates $\mathbf{H}_{t+1}$, closing the feedback loop.

\subsection{Energy-Structured Latent World Model}
Figure~\ref{fig:pipeline} demonstrates the ELWM overview from interaction latent encoding and pH-constrained motion prediction to PC-NTF.
\paragraph{Latent physical state.}
The ELWM world model encoder $E_\theta$ mappings the interaction history to the latent physical state:
\begin{equation}
(\mathbf{M}_t, \mathbf{Q}_t, \mathbf{V}_t) = E_\theta(\mathbf{H}_t^{\mathrm{phy}}),
\label{eq:physics-state}
\end{equation}
where $\mathbf{M}_t$ captures the slow physical regularity dynamics (payload, ground response, actuation delay), and $\mathbf{Q}_t, \mathbf{V}_t \in \mathbb{R}^{d_z/2}$ are latent generalized coordinates and velocities computed from relative motion features. The latent inertia $\mathbf{G}$ links velocity and momentum, and we parameterize it by Cholesky factor $\mathbf{c}_t$ with robot meta-info $\boldsymbol{\mu}_r$. 
\begin{equation}
\mathbf{G}_t
=
\mathbf{G}_\theta(\mathbf{M}_t,\boldsymbol{\mu}_r,\mathbf{c}_t)
\succ 0,
\qquad
\mathbf{P}_t=\mathbf{G}_t\mathbf{V}_t,
\label{eq:latent-momentum}
\end{equation}
We take $\mathbf{Z}_t=[\mathbf{Q}_t,\mathbf{P}_t]$ as the port-Hamiltonian latent phase state, and the pairing in Eq.~\eqref{eq:latent-momentum} supplies consistent velocity--momentum relation for the transition below.


\paragraph{Latent stored energy.}
We formalize the latent stored energy as
\begin{equation}
\Ham(\mathbf{Q}_t, \mathbf{P}_t; \mathbf{M}_t, \boldsymbol{\mu}_r)
=
\underbrace{\tfrac{1}{2} \mathbf{P}_t^\top \mathbf{G}_\theta^{-1} \mathbf{P}_t}_{\mathcal{K}_\theta}
+
\underbrace{U_\theta\!\big(\boldsymbol{\rho}(\mathbf{Q}_t), \mathbf{M}_t, \boldsymbol{\mu}_r\big)}_{\mathcal{U}_\theta},
\label{eq:hamiltonian}
\end{equation}
with kinetic energy $\mathcal{K}_\theta$ and potential energy $\mathcal{U}_\theta$. $\rho(\cdot)$ is a feature map restricted to relative quantities, which makes $\mathcal{U}_\theta$ invariant to global translations of $\mathbf{Q}_t$. Since kinetic parameterization gives $\partial \mathcal{K}_\theta / \partial \mathbf{P}_t = \mathbf{G}_t^{-1} \mathbf{P}_t = \mathbf{V}_t$, the Hamiltonian flow of $\mathbf{Q}_t$ coincides with the encoder's latent velocity. 

\paragraph{Port-Hamiltonian transition.}
Navigation robot exchanges energy with environments: actuators inject work, and the interaction dissipates it. Following port-Hamiltonian formalism \cite{zhong2019symplectic,desai2021port}, latent transition generated by the dissipation paradigm
\begin{equation}
\dot{\mathbf{Z}}_t
=
\big(\mathbf{J} - \mathbf{R}_\theta(\mathbf{Z}_t, \mathbf{M}_t, \boldsymbol{\mu}_r)\big) \nabla_\mathbf{Z} \Ham
+
B_\theta(\mathbf{Z}_t, \mathbf{M}_t, \boldsymbol{\mu}_r)\, \bm{u}_t,
\label{eq:ph-dynamics}
\end{equation}
with skew-symmetric operator $\mathbf{J} = \left[\begin{smallmatrix} 0 & \mathbf{I} \\ -\mathbf{I} & 0 \end{smallmatrix}\right]$, dissipation $\mathbf{R}_\theta \succeq 0$, control port $B_\theta$, and latent control input $\bm{u}_t = E_u(\bm{a}_t, \bm{o}_t, \boldsymbol{\mu}_r)$ mapping velocity and actuator commands.

The ELWM physical transition structural property follows directly, defining the port output $\bm{y}_t = B_\theta^\top \nabla_\mathbf{Z} \Ham$ and using $\mathbf{J}^\top = -\mathbf{J}$,
\begin{equation}
\frac{\text{d}\Ham}{\text{d}t}
=
\nabla_\mathbf{Z} \Ham^\top \dot{\mathbf{Z}}_t
=
-\,\nabla_\mathbf{Z} \Ham^\top \mathbf{R}_\theta \nabla_\mathbf{Z} \Ham
+
\bm{u}_t^\top \bm{y}_t.
\label{eq:energy-balance}
\end{equation}
Latent energy changes is the injected system latent work minus a non-negative dissipated power, and \emph{it is \textbf{conserved}} when $u_t = 0$ and $R_\theta = 0$.
Equation~\eqref{eq:energy-balance} holds as an identity of the vector field in Eq.~\eqref{eq:ph-dynamics} for energy balance; acceleration and friction are explained by the control port and dissipation terms. 
For momentum, the pH transition gives $\dot{\mathbf{P}}_t = -\partial \Ham / \partial \mathbf{Q}_t + [B_\theta \bm{u}_t - \mathbf{R}_\theta \nabla_\mathbf{Z} \Ham]_\mathbf{P}$; when the Hamiltonian is invariant along a latent direction and that direction is unforced and dissipation-free, the corresponding momentum component is constant.

Future predictions of latent states are produced by integrator $\Phi^{\mathrm{pH}}_{\Delta t}$. Since $G_\theta$ depends on $Q_t$, the energy in Eq.~\eqref{eq:hamiltonian} is non-separable, we use the implicit midpoint as discrete-gradient updates for dissipation. 

\paragraph{Training.}
In our design, physical regularities are carried by the transition structure, so the world model training task reduces to future prediction. Targets are encodings of future histories by the same encoder, $Z_{t+k}^{+} = E_\theta(H_{t+k}^{\mathrm{phy}})_Z$, and the objective is
\begin{equation}
\mathcal{L}_{\mathrm{ELWM}}
=
\frac{1}{K} \sum_{k=1}^{K} \frac{w_k}{d_z} \big\| \hat{Z}_{t+k} - Z_{t+k}^{+} \big\|_2^2
+
\lambda_{\mathrm{sig}}\, \mathcal{R}_{\mathrm{SIGReg}},
\label{eq:pwm-loss}
\end{equation}
the task-level future-latent prediction objective with representation-geometry regularizer: $\mathcal{R}_{\mathrm{SIGReg}}$ is the isotropic-Gaussian embedding regularizer of LeJEPA \cite{balestriero2025lejepa} applied to current, target, and rolled-out latents, which prevents collapse. 
Gradients propagate through the structure-preserving rollout into $\Ham, R_\theta, B_\theta, E_u$, we control the port-Hamiltonian mechanism constraints make effects that the model can reduce prediction error only after the physics. 

Energy residuals $\varepsilon_{\mathrm{energy}} = (\Delta \hat{\mathcal{E}}_t - \hat{W}_t)^2$ and symmetry-masked momentum residuals ar e logged for checkpoint selection; reinstating them in the loss is evaluated as an ablation. After convergence the world model exposes the physical context and world state for \system{} navigation planning.

\subsection{PC-NTF System Construction}
\begin{figure*}[t!]
    \centering
    \includegraphics[width=0.75\textwidth]{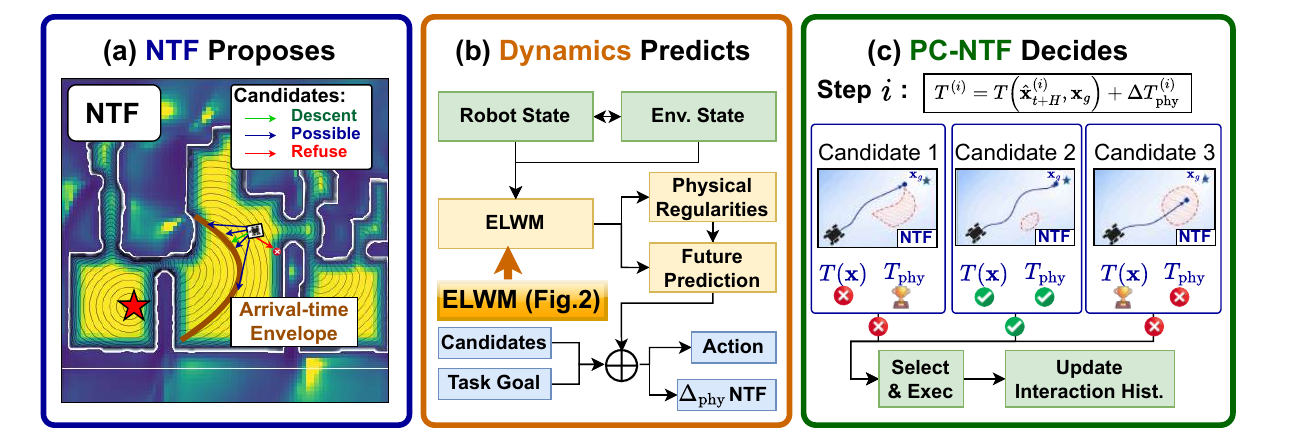}
    \caption{PC-NTF motion planning pipeline. (a) NTF proposes shared motion candidates -> (b) ELWM predicts their action-conditioned execution consequences -> (c) PC-NTF selects the candidate with the most efficient traversal delay.}
    \label{fig:framework}
    \vspace{-0.4cm}
\end{figure*}

PC-NTF implements the ELWM priors with neural time fields for complete navigation system, ELWM supplies implicit physics, and the Eikonal solver formalize the time fields initialization. Figure~\ref{fig:framework} illustrates how PC-NTF reranks NTF candidates using ELWM-predicted motion consequences in sequence. We build PC-NTF in three steps---latent physics decoder, condition time fields, and the generalized time-field problem\cite{liu2025physics}:
\begin{equation}
\underbrace{\sgeo(\mathbf{x})\, \|\nabla_\mathbf{x} T\|_2 = 1}_{\text{geometry Eikonal}}
\;\;\longrightarrow\;\;
\underbrace{\seff\big(\mathbf{x};\, \zeta_t(\mathbf{x}), \mu_r\big)\, \|\nabla_\mathbf{x} T\|_2 = 1}_{\text{physics-conditioned Eikonal}},
\label{eq:pc-ntf}
\end{equation}
where $\zeta_t(x) = E_{\mathrm{map}}(b_t, x)$ denotes map construction at the motion planning. Equation~\eqref{eq:pc-ntf} generalizes the solver objective, and additional dependency ($\zeta_t$, $\boldsymbol{\mu}_r$) is available before planning begins.

\paragraph{Physics latent decoder.}
Learning form the ELWM future prediction context, we take it as the action-conditioned consequences of motion planning. To better use the physical latent prediction for motion planning, we design the PC-NTF physics latent decoder which maps latent into traversal speed, and this decoder is trained by Huber loss from the physics-speed envelopes $\mathbf{s}^{*}_t$.
Given the interaction history $\mathbf{H}_t$ and a candidate command sequence $\mathbf{U}^{(i)}_t$, the decoder exposes planner-facing representation
\begin{equation}
\boldsymbol{\Gamma}^{(i)}_t
=
\mathcal{A}_{\mathrm{WM}}
\!\left(\mathbf{H}_t,\mathbf{U}^{(i)}_t\right)
=
\left(
\hat{\boldsymbol{\tau}}^{(i)}_t,\,
\hat{\mathbf{x}}^{(i)}_{t+H},\,
\Delta T^{(i)}_{\mathrm{WM}},\,
\boldsymbol{\rho}^{(i)}_t
\right),
\label{eq:wm-planning-interface}
\end{equation}
where $\hat{\boldsymbol{\tau}}^{(i)}_t$ is the decoded observable motion, $\hat{\mathbf{x}}^{(i)}_{t+H}$ is its terminal state, $\Delta T^{(i)}_{\mathrm{WM}}\geq 0$ is traversal delay, and $\boldsymbol{\rho}^{(i)}_t$ summarizes physical predicted tracking, safety, and control consequences.

\paragraph{NTF Construction and Physics Condition.}
NTF represents scene by a continuous arrival-time field $T(\mathbf{x},\mathbf{g})$, which estimates the remaining travel time from a location $\mathbf{x}$ to the goal $\mathbf{x}_g$. Its geometry-derived speed $S(\mathbf{x})$ and arrival time follow the Eikonal relation
\begin{equation}
S(\mathbf{x}) \left\lVert \nabla T(\mathbf{x},\mathbf{x}_g) \right\rVert_2 =1,
\qquad
T(\mathbf{g},\mathbf{g})=0 .
\label{eq:antf-eikonal}
\end{equation}
Time field is parameterized by PINN neural network and fitted from location--goal pairs using the Eikonal equation and the zero-time boundary condition at the goal. As new geometric observations become available, Active NTF updates the clearance-dependent speed supervision and the corresponding arrival-time representation without requiring expert trajectories \cite{ni2022ntfields}. 
Regions with greater obstacle clearance admit faster propagation and therefore smaller arrival time, while following the negative field gradient provides a goal-directed motion cue. This field captures scene geometry efficiently in navigation, but it cannot by itself distinguish physical changes in dissipation, actuation, or motion response.

PC-NTF complements the geometric field with action-conditioned predictions from the ELWM. The NTF descent direction first defines a nominal motion, from which the planner forms a small set of feasible alternatives.
ELWM rolls out every alternative from the same interaction history and decodes its predicted pose and velocity. These predictions $\mathcal{P}$ go through ``pH'' and decoded by the physics latent decoder. determine two quantities that are directly meaningful to the time field: the location that the robot is expected to reach and any additional time caused by slower-than-commanded progress. For candidate $i$, PC-NTF evaluates
\begin{equation}
T^{(i)} = T\!\left( \hat{\mathbf{x}}^{(i)}_{t+H},\mathbf{x}_g \right) + \Delta T^{(i)}_{\mathrm{phys}}.
\label{eq:pcntf-time}
\end{equation}
Neural time fields solve the Eikonal equation, $T\!\left( \hat{\mathbf{x}}^{(i)}_{t+H},\mathbf{x}_g \right)$ is the geometry time remaining from the predicted terminal location. ELWM provide the $T^{(i)}_{\mathrm{phys}}$ after decoder, representing the physical regularities and predicted motion by the world model along with the navigation progress. 
Consequently, a geometrically attractive motion receives a larger cost when the current physical regime makes it difficult to realize. PC-NTF therefore physically conditions the use of the neural time field without treating the latent state, mechanical energy, or dissipation matrix as an Eikonal speed.

\paragraph{Motion-planning and navigation system.}
At every replanning step, the system follows one execution order: query the NTF field, generate shared feasible candidates, predict their motion with the world model, and evaluate them using Eq.~\eqref{eq:pcntf-time}.
Candidates with poor tracking, speed-limit violations, high uncertainty, or abrupt control changes are penalized, and the first action of the lowest-cost feasible candidate is executed. The resulting pose, velocity, and executed action are then appended to the interaction history before the next planning step. The ELWM and NTF are trained separately from interaction trajectories and scene geometry, respectively, and remain frozen during system evaluation. This construction preserves NTF as the source of global geometric guidance while using the world model to supply the physical consequences that a geometry-only time field cannot observe.

Equation~\eqref{eq:pc-ntf} is also the mechanism by which the world model influences planning: when the readout reports reduced capability---higher dissipation, heavier payload---$\seff$ drops, the Eikonal constraint forces larger time gradients, the same Euclidean distance costs more arrival time, and planned paths bend away from physically slow regions without any explicit penalty. Motion generation follows the solved field:
\begin{equation}
\begin{aligned}
\mathbf{S}^{\mathrm{exec}}_t = \Pi_{C_t}\big[ - \seff(\mathbf{x})\, \frac{\nabla T(\mathbf{x}; \mathbf{x}_g)}{\|\nabla T(\mathbf{x}; \mathbf{x}_g)\|_2 + \epsilon} \big],
\label{eq:velocity-field}
\end{aligned}
\end{equation}
where the implied speed $\hat{s}_t = 1/(\|\nabla_x T_t\|_2 + \epsilon) \approx \seff$ serves as a built-in diagnostic of how faithfully the solver absorbed the physical prior, and $\Pi_{C_t}$ is a deterministic projection enforcing occupancy boundaries, unknown-space rules, velocity and acceleration clipping, and emergency stops. 

\section{Experiments}



\subsection{Experimental Setup}

\noindent\textbf{Environments.} (i) iGibson indoor scenes \cite{shen2020igibson} with cluttered multi-room layouts; (ii) held-out environments sampled from HM3D \cite{ramakrishnan2021habitat} for topological complexity, formalize into iGibson format; (iii) a terrain-variation suite with heterogeneous ground surfaces (carpet, tile, gravel) and payload changes that alter interaction dynamics. Train/test splits are by scene, surface, and robot meta-configuration; adjacent frames have bounding split.


\noindent\textbf{Baselines.} The main comparison includes Active Neural Time Fields\cite{liu2025physics}, Generic world models (the capacity matched generative action-consistent world model)\cite{chen2026lawam} with the MPC budget, and the complete ELWM+PC-NTF system. 

\noindent\textbf{Comparison axes.}
We organize the evaluation following two axes. 
At the \emph{world-model axis}, we ask whether imposing an energy-structured transition improves action-conditioned prediction over a capacity-matched unconstrained latent world model. ELWM is compared with a capacity-matched Generic WM~\cite{chen2026lawam}. 
At the \emph{planning system axis}, we ask how these predictions change the motions selected from the same ANTF candidate set, PC-NTF is compared with geometry ANTF~\cite{liu2025physics}.

\noindent\textbf{Metrics.}
The world-model axis uses 0.8-s motion-prediction normalized root mean square error (NRMSE@0.8s; lower is better), pose error, and twist error. The planning axis uses success rate, success weighted by path length (SPL), physical collision rate, and Eikonal residual.

\subsection{Experimental Results}

\paragraph{Physical Consistent World Model Prediction.}
Table~\ref{tab:wm-physical-consistency} isolates the predictive contribution of the energy-structured transition. ELWM reduces observable-motion NRMSE from \(0.36\) to \(0.29\) relative to Generic WM, an \(19.4\%\) reduction.  Pose and twist errors respectively decrease by \(10.8\%\) and \(1.8\%\). Along with the experiment, a larger action-intervention ratio (\(1.193\) to \(1.009\)) shows that the ELWM rollout depends more strongly on the executed control.  Also, there is a smaller positive difference in slow-\(M\) intervention (\(1.062\) to \(1.033\)), showing that \(M_t\) has the potential to identify reusable physical regularities.

\begin{table}[t]
    \centering
    \footnotesize
    \setlength{\tabcolsep}{3.2pt}
    \renewcommand{\arraystretch}{1.12}
    \begin{tabular}{@{}lcccc@{}}
        \toprule
        \textbf{Method}
        & \makecell{\textbf{NRMSE}\\[-1pt]\textbf{@0.8s}$\downarrow$}
        & \textbf{Pose}$\downarrow$
        & \textbf{Twist}$\downarrow$
        & \textbf{Action int.}$\uparrow$ \\
        \midrule

        \makecell[l]{Generic WM}
        & 0.36
        & 0.749
        & 0.489
        & 1.009 \\

        \addlinespace[2pt]
        \textbf{\makecell[l]{ELWM}}
        & \textbf{0.29}
        & \textbf{0.669}
        & \textbf{0.480}
        & \textbf{1.193} \\

        \bottomrule
    \end{tabular}
    \caption{\textbf{World-model physical consistency analysis.} Prediction errors are evaluated on paired cases under identical observations, action constraints, and candidate motions in indoor scenes.}
    \label{tab:wm-physical-consistency}
    \vspace{-0.2cm}
\end{table}

Physical world model analysis includes energy consistency, our trained ELWM checkpoint attains an energy-balance residual of \(1.88\times10^{-7}\), and its predicted dissipated power is nonnegative on all evaluated rollout steps, verified that the numerical transition respects the imposed dissipative pH identity and ELWM energy-structured model.

\begin{table}[t]
\centering
\setlength{\tabcolsep}{3.0pt}
\renewcommand{\arraystretch}{1.16}
\begin{tabular}{@{}llcc@{}}
\toprule
\textbf{Metrics}
& \textbf{Setting}
& \textbf{Regime}$\downarrow$
& \textbf{Scene}$\downarrow$ \\
\midrule
\multirow[c]{2}{*}{%
  \makecell[l]{\textbf{Motion}\\NRMSE}}
& Random init.
& 7.753
& 7.699 \\
& \cellcolor{cyan!9}\textbf{ELWM}
& \cellcolor{cyan!9}
  \makecell[c]{%
    \textbf{0.705}\\[-0.5ex]
    {\scriptsize\textcolor{teal!70!black}{$\downarrow\,90.9\%$}}}
& \cellcolor{cyan!9}
  \makecell[c]{%
    \textbf{0.697}\\[-0.5ex]
    {\scriptsize\textcolor{teal!70!black}{$\downarrow\,91.0\%$}}} \\
\midrule
\multirow[c]{2}{*}{%
  \makecell[l]{\textbf{Latent}\\MSE}}
& Random init.
& 0.1695
& 0.1656 \\
& \cellcolor{cyan!9}\textbf{ELWM}
& \cellcolor{cyan!9}
  \makecell[c]{%
    \textbf{0.0856}\\[-0.5ex]
    {\scriptsize\textcolor{teal!70!black}{$\downarrow\,49.5\%$}}}
& \cellcolor{cyan!9}
  \makecell[c]{%
    \textbf{0.0835}\\[-0.5ex]
    {\scriptsize\textcolor{teal!70!black}{$\downarrow\,49.6\%$}}} \\
\bottomrule
\end{tabular}
\caption{\textbf{Comparison with random initialization}, showing ELWM learning effects and generalization on held-out cases.}
\label{tab:physical-main}
\vspace{-0.2cm}
\end{table}
ELWM substantially improves prediction on both held-out axes, motion NRMSE decreases from 7.753 to \textbf{0.705} on unseen physical regimes and from 7.699 to \textbf{0.697} on unseen scenes, corresponding to relative reductions of 90.9\% and 91.0\%. Latent MSE is also reduced by 49.5\% and 49.6\%. The closely matched reductions show that ELWM training process effectively improves both held-out tests on hold out.

\begin{figure}[t]
    \centering
    \includegraphics[width=0.8\columnwidth]{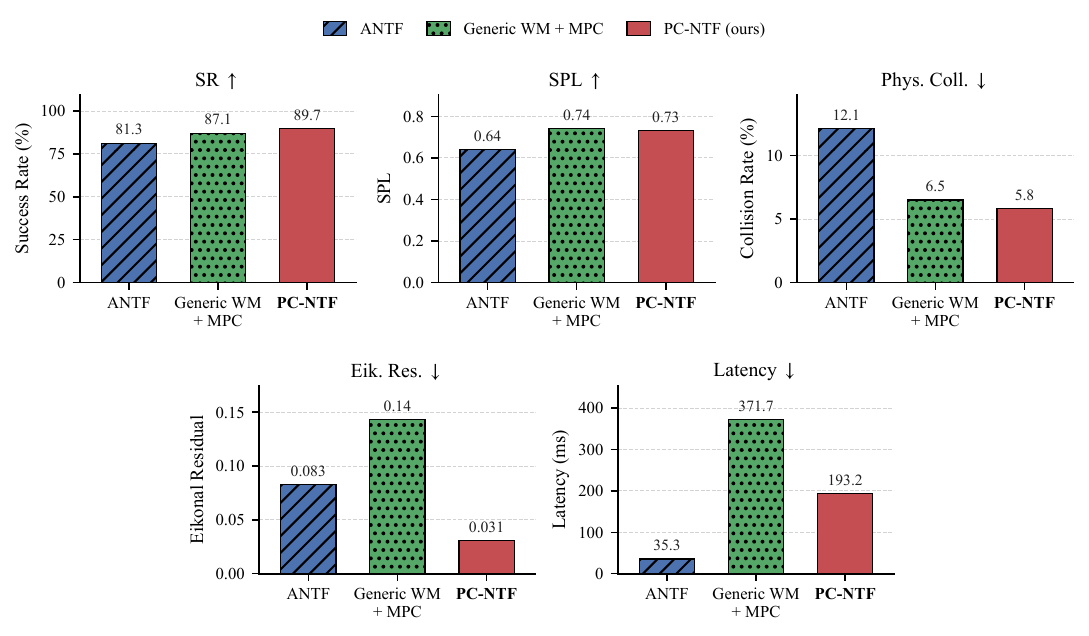}
    \caption{\textbf{Motion-planning and navigation results comparison}, PC-NTF indicates ELWM+PC-NTF system.}
    \label{fig:planning-main}
    \vspace{-0.3cm}
\end{figure}

\paragraph{Motion Planning and Navigation.}
Table~\ref{tab:planning-main} shows the motion planning systems comparison, and Figure~\ref{fig:planning-main} visualizes the matched planning comparison. For efficient and physical motion planning, PC-NTF is intended to improve physics-awared navigation because ELWM supplies a physics-conditioned estimation of how the current system will respond to navigation interaction based on learned physical regularities. 
This prior modifies the effective propagation cost and penalizes candidates that are trajectory valid but physically slow or difficult to execute, PC-NTF formalize them into Neural Time Fields for motion planning. Compared with ANTF, PC-NTF improves SR from 81.3\% to \textbf{89.7\%}, SPL from 0.64 to \textbf{0.73}, reduce Eikonal residual from 0.083 to \textbf{0.031}. ELWM physics prior bring more latency than origin ANTF, while we expose that PC-NTF latency is better than MPC control port (371.7 to \textbf{193.2}).

\begin{table}[t]
    \centering
    \footnotesize
    \setlength{\tabcolsep}{2.3pt}
    \renewcommand{\arraystretch}{1.12}
    \begin{tabular}{@{}lccccc@{}}
        \toprule
        \textbf{Method}
        & \makecell{\textbf{SR}\\[-1pt]\textbf{(\%)}$\uparrow$}
        & \textbf{SPL}$\uparrow$
        & \makecell{\textbf{Phys. Coll.}\\[-1pt]\textbf{(\%)}$\downarrow$}
        & \textbf{Eik.}$\downarrow$
        & \makecell{\textbf{Latency}\\[-1pt]\textbf{(ms)}$\downarrow$} \\
        \midrule

        ANTF
        & 81.3
        & 0.64
        & 14.4
        & 0.083
        & \textbf{35.3} \\

        \makecell[l]{Generic WM\\+ MPC}
        & 87.1
        & \textbf{0.74}
        & 6.5
        & 0.143
        & 371.7 \\

        \addlinespace[2pt]
        \textbf{\makecell[l]{ELWM\\+ PC-NTF}}
        & \textbf{89.7}
        & 0.73
        & \textbf{5.8}
        & \textbf{0.031}
        & 193.2 \\

        \bottomrule
    \end{tabular}
    \caption{\textbf{Motion-planning and navigation performance.} All methods use identical maps, observations, action constraints, candidate sets, and planning budgets.}
    \label{tab:planning-main}
    \vspace{-0.3cm}
\end{table}

We stress-test ELWM along two complementary held-out axes to test the learned transition reliability beyond the training configurations. The \emph{held-out-regime} split changes the robot--environment interaction conditions, whereas the \emph{held-out-scene} split changes the surrounding geometry. Both evaluations use the same 0.8\,s recorded-action rollout horizon. A randomly initialized ELWM with the same architecture serves only as a sanity reference, so Table~\ref{tab:physical-main} measures whether training yields action-conditioned prediction across the two held-out splits.

\subsection{Scalability and Ablation Studies}

\begin{figure}[t]
    \centering
    \includegraphics[width=0.8\columnwidth]{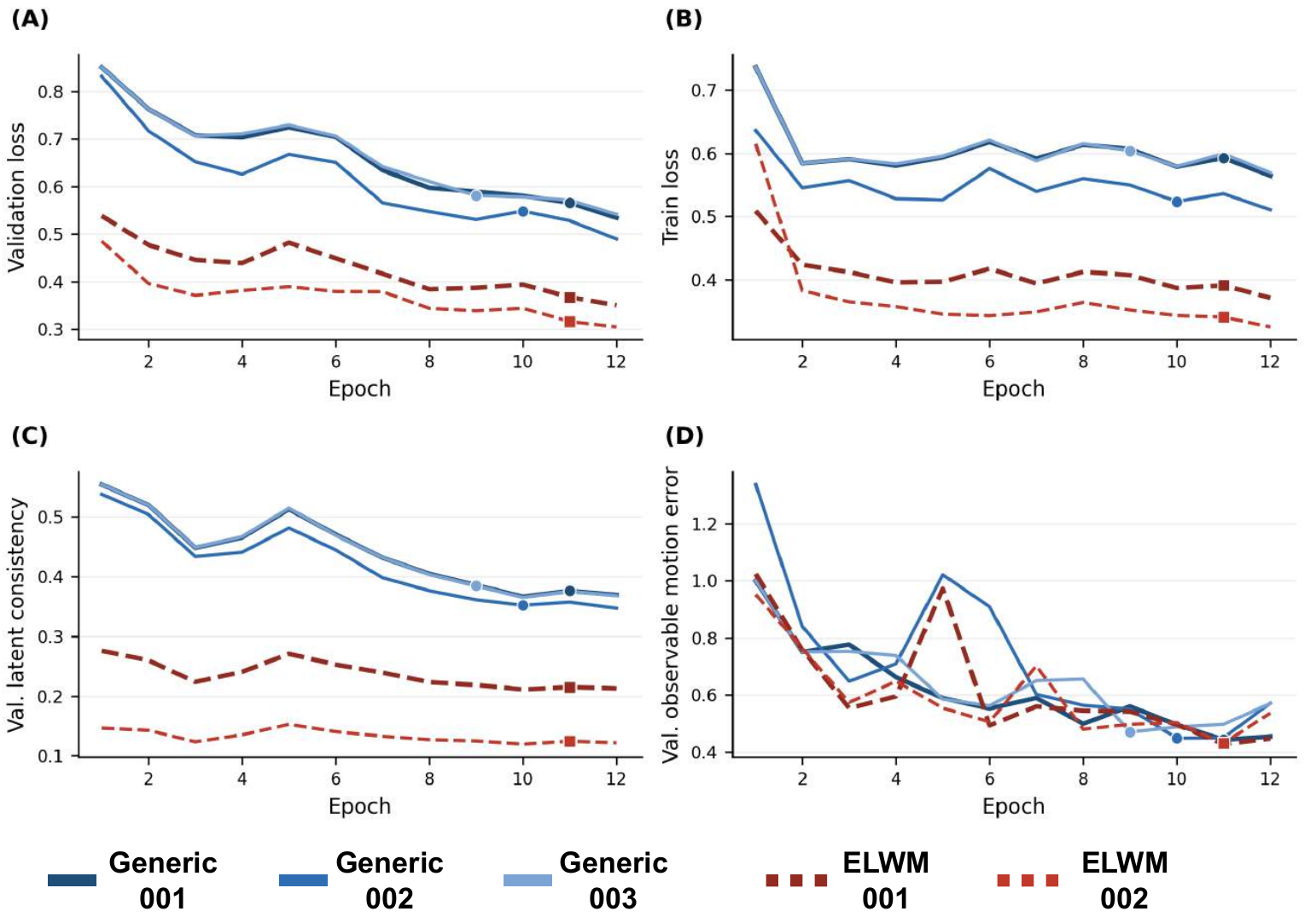}
    \caption{\textbf{Convergence analysis across scalability training configurations.} Compared in terms of (A) validation loss, (B) training loss, (C) latent-consistency error, and (D) observable-motion error, demonstrating ELWM stable learning beyond the hyperparameter setting.}
    \label{fig:parameter-convergence}
    \vspace{-0.2cm}
\end{figure}

\noindent \textbf{Scalability.}
We investigate whether the learning behavior of ELWM depends on tuned optimization setting. Specifically, ELWM and the Generic WM baseline are trained under a matched protocol while varying combinations of the learning rate, weight decay, and the SIGReg coefficient $\lambda_{\mathrm{sig}}$; all other data, optimization, and evaluation settings are kept fixed. As shown in Figure~\ref{fig:parameter-convergence}, ELWM consistently reaches a lower-loss region within the first few epochs and maintains lower training loss, validation loss, and future-latent consistency error across the tested configurations. The small variation between ELWM runs further indicates that this behavior is not tied to a single parameter combination. Meanwhile, the two model families eventually reach a comparable range of observable-motion error, showing that the improvement in latent prediction is not obtained by sacrificing motion-level accuracy. These results demonstrate that the physics-structured transition provides stable and parameter-robust learning across the evaluated optimization configurations.

\noindent\textbf{Ablations.}
Table~\ref{tab:planning-ablation} isolates where the PC-NTF gain comes from. The ANTF row measures geometry-only propagation. Replacing pH dynamics with a capacity-matched generic transition tests the ELWM structure; neutralizing the ELWM condition tests whether learned physics actually reaches the time field; and the scalar-only variant separates global speed calibration from spatial capability and candidate-specific traversal delay. 

\begin{table}[t]
    \centering
    \small
    \resizebox{\columnwidth}{!}{%
    \begin{tabular}{lcccc}
        \toprule
        Variant & NRMSE$\downarrow$ & SPL$\uparrow$ & Phys. Coll.$\downarrow$ & Latency (ms)$\downarrow$ \\
        \midrule
        ANTF & 1.269 & 0.597 & 14.35\% & \textbf{40} \\
        PC-NTF w/o physical condition & 2.347 & 0.483 & 18.75\% & 279 \\
        PC-NTF w/ shuffled $M_t$ & 1.036 & 0.580 & 22.11\% & 266 \\
        PC-NTF w/ generic WM & 1.015 & 0.654 & 6.54\% & 479 \\
        \textbf{ELWM + PC-NTF} & \textbf{1.012} & \textbf{0.661} & \textbf{5.79\%} & 278 \\
        \bottomrule
    \end{tabular}%
    }
    \caption{Ablation study of PC-NTF. Prediction horizon and latency is measured per replanning cycle.}
    \label{tab:planning-ablation}
    \vspace{-0.2cm}
\end{table}

\section{Conclusion}

In this work, we investigated how interaction-derived physical knowledge can be represented by latent world models and translated into executable guidance for motion planning in unknown environments. We propose the Energy-Structured Latent World Model (\textbf{ELWM}) that structures action-conditioned prediction through port-Hamiltonian dynamics with explicit dissipation and control ports.
For reusable physical regularities learning, we introduced Physics-Conditioned Neural Time Fields (\textbf{PC-NTF}) as the navigating motion system realization. PC-NTF connects predicted motion consequences to the global arrival-time representation of Neural Time Fields through a physically interpretable time-to-go objective. This construction preserves the efficiency and geometric guidance of neural time fields while allowing changes in robot--environment dynamics to inform motion selection. Our two-level evaluation analyses physically consistent world-model prediction and closed-loop motion planning, providing evaluation for ELWM physical consistent latent prediction quality and system-level planning benefit. The physical core evaluation also supports ELWM prediction and physical consistency on held-out scopes; paired navigation evaluation provides the corresponding benefit of PC-NTF motion planning.

\noindent\textbf{Future Work.} We are delighted to extend PC-NTF from planar navigation to anisotropic, state-space, and contact-rich motion planning, and to evaluate transfer across broader robot configurations and real-world environments. 

\bibliography{references}

@article{ha2018recurrent,
  title={Recurrent world models facilitate policy evolution},
  author={Ha, David and Schmidhuber, J{\"u}rgen},
  journal={Advances in neural information processing systems},
  volume={31},
  year={2018}
}

@inproceedings{hafner2019learning,
  title={Learning latent dynamics for planning from pixels},
  author={Hafner, Danijar and Lillicrap, Timothy and Fischer, Ian and Villegas, Ruben and Ha, David and Lee, Honglak and Davidson, James},
  booktitle={International conference on machine learning},
  pages={2555--2565},
  year={2019},
  organization={PMLR}
}

@article{hafner2019dream,
  title={Dream to control: Learning behaviors by latent imagination},
  author={Hafner, Danijar and Lillicrap, Timothy and Ba, Jimmy and Norouzi, Mohammad},
  journal={arXiv preprint arXiv:1912.01603},
  year={2019}
}

@article{hafner2023mastering,
  title={Mastering diverse domains through world models},
  author={Hafner, Danijar and Pasukonis, Jurgis and Ba, Jimmy and Lillicrap, Timothy},
  journal={arXiv preprint arXiv:2301.04104},
  year={2023}
}

@article{balestriero2025lejepa,
  title={Lejepa: Provable and scalable self-supervised learning without the heuristics},
  author={Balestriero, Randall and LeCun, Yann},
  journal={arXiv preprint arXiv:2511.08544},
  year={2025}
}

@inproceedings{zhou2025dino,
  title={DINO-WM: World Models on Pre-trained Visual Features enable Zero-shot Planning},
  author={Zhou, Gaoyue and Pan, Hengkai and Lecun, Yann and Pinto, Lerrel},
  booktitle={International Conference on Machine Learning},
  pages={79115--79135},
  year={2025},
  organization={PMLR}
}

@inproceedings{bar2025navigation,
  title={Navigation World Models},
  author={Bar, Amir and Zhou, Gaoyue and Tran, Danny and Darrell, Trevor and LeCun, Yann},
  booktitle={2025 IEEE/CVF Conference on Computer Vision and Pattern Recognition (CVPR)},
  pages={15791--15801},
  year={2025},
  organization={IEEE}
}

@article{chen2026lawam,
  title={Lawam: Latent world action models for efficient dynamics-aware robot policies},
  author={Chen, Jialei and Wang, Kai and Chen, Kang and Chen, Shuaihang and Gao, Feng and Tang, Wenhao and Li, Zhiyuan and Liu, Weilin and Yao, Zhuyu and Li, Boxun and others},
  journal={arXiv preprint arXiv:2606.15768},
  year={2026}
}

@article{shang2026roboscape,
  title={Roboscape: Physics-informed embodied world model},
  author={Shang, Yu and Zhang, Xin and Tang, Yinzhou and Jin, Lei and Gao, Chen and Wu, Wei and Li, Yong},
  journal={Advances in Neural Information Processing Systems},
  volume={38},
  pages={63674--63698},
  year={2026}
}

@article{greydanus2019hamiltonian,
  title={Hamiltonian neural networks},
  author={Greydanus, Samuel and Dzamba, Misko and Yosinski, Jason},
  journal={Advances in neural information processing systems},
  volume={32},
  year={2019}
}

@article{desai2021port,
  title={Port-Hamiltonian neural networks for learning explicit time-dependent dynamical systems},
  author={Desai, Shaan and Mattheakis, Marios and Sondak, David and Protopapas, Pavlos and Roberts, Stephen},
  journal={arXiv preprint arXiv:2107.08024},
  year={2021}
}

@article{rao2026skyjepa,
  title={Skyjepa: Learning long-horizon world models for zero-shot sim-to-real control of quadrotors},
  author={Rao, Pratyaksh and Zhang, Wancong and Balestriero, Randall and LeCun, Yann and Loianno, Giuseppe},
  journal={arXiv preprint arXiv:2606.23444},
  year={2026}
}

@article{raissi2018physics,
  title={Physics-informed neural networks: A deep learning framework for solving forward and inverse problems involving nonlinear partial differential equations},
  author={Raissi, Maziar and Perdikaris, Paris and Karniadakis, George Em},
  journal={Journal of Computational physics},
  volume={378},
  number={C},
  year={2018},
  publisher={Univ. of Pennsylvania, Philadelphia, PA (United States)}
}

@article{sethian1996fast,
  title={A fast marching level set method for monotonically advancing fronts.},
  author={Sethian, James A},
  journal={proceedings of the National Academy of Sciences},
  volume={93},
  number={4},
  pages={1591--1595},
  year={1996}
}

@article{karaman2011sampling,
  title={Sampling-based algorithms for optimal motion planning},
  author={Karaman, Sertac and Frazzoli, Emilio},
  journal={The international journal of robotics research},
  volume={30},
  number={7},
  pages={846--894},
  year={2011},
  publisher={Sage Publications Sage UK: London, England}
}

@article{qureshi2020motion,
  title={Motion planning networks: Bridging the gap between learning-based and classical motion planners},
  author={Qureshi, Ahmed Hussain and Miao, Yinglong and Simeonov, Anthony and Yip, Michael C},
  journal={IEEE Transactions on Robotics},
  volume={37},
  number={1},
  pages={48--66},
  year={2020},
  publisher={IEEE}
}

@article{smith2020eikonet,
  title={Eikonet: Solving the eikonal equation with deep neural networks},
  author={Smith, Jonathan D and Azizzadenesheli, Kamyar and Ross, Zachary E},
  journal={IEEE Transactions on Geoscience and Remote Sensing},
  volume={59},
  number={12},
  pages={10685--10696},
  year={2020},
  publisher={IEEE}
}

@article{bin2021pinneik,
  title={PINNeik: Eikonal solution using physics-informed neural networks},
  author={Bin Waheed, Umair and Haghighat, Ehsan and Alkhalifah, Tariq and Song, Chao and Hao, Qi},
  journal={Computers \& Geosciences},
  volume={155},
  pages={104833},
  year={2021},
  publisher={Elsevier}
}

@article{ni2022ntfields,
  title={Ntfields: Neural time fields for physics-informed robot motion planning},
  author={Ni, Ruiqi and Qureshi, Ahmed H},
  journal={arXiv preprint arXiv:2210.00120},
  year={2022}
}

@article{liu2025physics,
  title={Physics-informed neural mapping and motion planning in unknown environments},
  author={Liu, Yuchen and Ni, Ruiqi and Qureshi, Ahmed H},
  journal={IEEE Transactions on Robotics},
  year={2025},
  publisher={IEEE}
}

@article{zhang2026physics,
  title={Physics-Informed Eikonal Caging for Whole-Arm Manipulation Planning},
  author={Zhang, Yan and Li, Yiming and Dong, Yifei and Pokorny, Florian T and Calinon, Sylvain},
  journal={arXiv preprint arXiv:2606.22143},
  year={2026}
}

@article{kahn2021badgr,
  title={Badgr: An autonomous self-supervised learning-based navigation system},
  author={Kahn, Gregory and Abbeel, Pieter and Levine, Sergey},
  journal={IEEE Robotics and Automation Letters},
  volume={6},
  number={2},
  pages={1312--1319},
  year={2021},
  publisher={IEEE}
}

@article{gasparino2022wayfast,
  title={Wayfast: Navigation with predictive traversability in the field},
  author={Gasparino, Mateus V and Sivakumar, Arun N and Liu, Yixiao and Velasquez, Andres EB and Higuti, Vitor AH and Rogers, John and Tran, Huy and Chowdhary, Girish},
  journal={IEEE Robotics and Automation Letters},
  volume={7},
  number={4},
  pages={10651--10658},
  year={2022},
  publisher={IEEE}
}

@article{shen2020igibson,
  title={igibson 1.0: A simulation environment for interactive tasks in large realistic scenes},
  author={Shen, Bokui and Xia, Fei and Li, Chengshu and Mart{\'\i}n-Mart{\'\i}n, Roberto and Fan, Linxi and Wang, Guanzhi and P{\'e}rez-D'Arpino, Claudia and Buch, Shyamal and Srivastava, Sanjana and Tchapmi, Lyne P and others},
  journal={arXiv preprint arXiv:2012.02924},
  year={2020}
}

@article{ramakrishnan2021habitat,
  title={Habitat-matterport 3d dataset (hm3d): 1000 large-scale 3d environments for embodied ai},
  author={Ramakrishnan, Santhosh K and Gokaslan, Aaron and Wijmans, Erik and Maksymets, Oleksandr and Clegg, Alex and Turner, John and Undersander, Eric and Galuba, Wojciech and Westbury, Andrew and Chang, Angel X and others},
  journal={arXiv preprint arXiv:2109.08238},
  year={2021}
}

@inproceedings{lions2006optimal,
  title={Optimal control and viscosity solutions},
  author={Lions, PL},
  booktitle={Recent Mathematical Methods in Dynamic Programming: Proceedings of the Conference held in Rome, Italy, March 26--28, 1984},
  pages={94--112},
  year={2006},
  organization={Springer}
}

@article{chen2025vl,
  title={Vl-jepa: Joint embedding predictive architecture for vision-language},
  author={Chen, Delong and Shukor, Mustafa and Moutakanni, Theo and Chung, Willy and Yu, Jade and Kasarla, Tejaswi and Bang, Yejin and Bolourchi, Allen and LeCun, Yann and Fung, Pascale},
  journal={arXiv preprint arXiv:2512.10942},
  year={2025}
}

@article{nie2026phys,
  title={Phys-JEPA: Physics-Informed Latent World Models for Multivariate Time-Series Forecasting},
  author={Nie, Weizhi and Liu, Weichao and Guo, Honglin and Su, Yuting},
  journal={arXiv preprint arXiv:2606.16076},
  year={2026}
}

@article{scholkopf2021toward,
  title={Toward causal representation learning},
  author={Sch{\"o}lkopf, Bernhard and Locatello, Francesco and Bauer, Stefan and Ke, Nan Rosemary and Kalchbrenner, Nal and Goyal, Anirudh and Bengio, Yoshua},
  journal={Proceedings of the IEEE},
  volume={109},
  number={5},
  pages={612--634},
  year={2021},
  publisher={IEEE}
}

@article{li2025pin,
  title={Pin-wm: Learning physics-informed world models for non-prehensile manipulation},
  author={Li, Wenxuan and Zhao, Hang and Yu, Zhiyuan and Du, Yu and Zou, Qin and Hu, Ruizhen and Xu, Kai},
  journal={arXiv preprint arXiv:2504.16693},
  year={2025}
}

@inproceedings{park2019deepsdf,
  title={Deepsdf: Learning continuous signed distance functions for shape representation},
  author={Park, Jeong Joon and Florence, Peter and Straub, Julian and Newcombe, Richard and Lovegrove, Steven},
  booktitle={Proceedings of the IEEE/CVF conference on computer vision and pattern recognition},
  pages={165--174},
  year={2019}
}

@article{you2023generative,
  title={Generative neural fields by mixtures of neural implicit functions},
  author={You, Tackgeun and Kim, Mijeong and Kim, Jungtaek and Han, Bohyung},
  journal={Advances in Neural Information Processing Systems},
  volume={36},
  pages={20352--20370},
  year={2023}
}

@article{du2011robot,
  title={Robot motion planning in dynamic, uncertain environments},
  author={Du Toit, Noel E and Burdick, Joel W},
  journal={IEEE Transactions on Robotics},
  volume={28},
  number={1},
  pages={101--115},
  year={2011},
  publisher={IEEE}
}

@article{lecun2022path,
  title={A path towards autonomous machine intelligence version 0.9. 2, 2022-06-27},
  author={LeCun, Yann and others},
  journal={Open Review},
  volume={62},
  number={1},
  pages={1--62},
  year={2022}
}

@article{zhong2025multi,
  title={Multi-objective trajectory planning for flexible spacecraft via physics-informed neural network},
  author={Zhong, Xilin and Liu, Shujie and Chen, Ti and Hu, Haiyan},
  journal={Aerospace Science and Technology},
  pages={110710},
  year={2025},
  publisher={Elsevier}
}

@article{luan2026ph,
  title={PH-Dreamer: A Physics-Driven World Model via Port-Hamiltonian Generative Dynamics},
  author={Luan, Xueyu and Shi, Chenwei},
  journal={arXiv preprint arXiv:2605.18303},
  year={2026}
}

@article{zhong2019symplectic,
  title={Symplectic ode-net: Learning hamiltonian dynamics with control},
  author={Zhong, Yaofeng Desmond and Dey, Biswadip and Chakraborty, Amit},
  journal={arXiv preprint arXiv:1909.12077},
  year={2019}
}

@inproceedings{troch2025action,
  title={Action-Conditioned Hamiltonian Generative Networks (AC-HGN) for Supervised and Reinforcement Learning},
  author={Troch, Arne and Mets, Kevin and Mercelis, Siegfried},
  booktitle={7th Annual Learning for Dynamics \& Control Conference, 04-06 June, 2025, Ann Arbor, Michigan, USA},
  pages={310--322},
  year={2025}
}

@article{hsu2002randomized,
  title={Randomized kinodynamic motion planning with moving obstacles},
  author={Hsu, David and Kindel, Robert and Latombe, Jean-Claude and Rock, Stephen},
  journal={The International Journal of Robotics Research},
  volume={21},
  number={3},
  pages={233--255},
  year={2002},
  publisher={SAGE Publications Sage UK: London, England}
}

@article{liu2025aligning,
  title={Aligning cyber space with physical world: A comprehensive survey on embodied ai},
  author={Liu, Yang and Chen, Weixing and Bai, Yongjie and Liang, Xiaodan and Li, Guanbin and Gao, Wen and Lin, Liang},
  journal={IEEE/ASME Transactions on Mechatronics},
  year={2025},
  publisher={IEEE}
}

@inproceedings{pham2013kinodynamic,
  title={Kinodynamic Planning in the Configuration Space via Admissible Velocity Propagation.},
  author={Pham, Quang-Cuong and Caron, St{\'e}phane and Nakamura, Yoshihiko},
  booktitle={Robotics: Science and Systems},
  volume={32},
  year={2013}
}

@article{bohg2017interactive,
  title={Interactive perception: Leveraging action in perception and perception in action},
  author={Bohg, Jeannette and Hausman, Karol and Sankaran, Bharath and Brock, Oliver and Kragic, Danica and Schaal, Stefan and Sukhatme, Gaurav S},
  journal={IEEE Transactions on Robotics},
  volume={33},
  number={6},
  pages={1273--1291},
  year={2017},
  publisher={IEEE}
}

@article{huang2026h,
  title={H-wm: Robotic task and motion planning guided by hierarchical world model},
  author={Huang, Jinbang and Chen, Wenyuan and Li, Zhiyuan and Pang, Oscar and Hu, Xiao and Zhang, Lingfeng and Hu, Yuanzhao and Zhang, Zhanguang and Coates, Mark and Cao, Tongtong and others},
  journal={arXiv preprint arXiv:2602.11291},
  year={2026}
}

@inproceedings{liu2026siam,
  title={SIAM: Towards Generalizable Articulated Object Modeling via Single Robot-Object Interaction},
  author={Liu, Yuyan and Zhang, Li and Wu, Di and Zhang, Yan and Huang, Anran and Wang, Zhi and Liu, Liu and Guo, Dan},
  booktitle={Proceedings of the AAAI Conference on Artificial Intelligence},
  volume={40},
  pages={18478--18486},
  year={2026}
}

@article{maes2026leworldmodel,
  title={Leworldmodel: Stable end-to-end joint-embedding predictive architecture from pixels},
  author={Maes, Lucas and Lidec, Quentin Le and Scieur, Damien and LeCun, Yann and Balestriero, Randall},
  journal={arXiv preprint arXiv:2603.19312},
  year={2026}
}

@article{lutter2019deep,
  title={Deep lagrangian networks: Using physics as model prior for deep learning},
  author={Lutter, Michael and Ritter, Christian and Peters, Jan},
  journal={arXiv preprint arXiv:1907.04490},
  year={2019}
}

@article{gong2026csympnet,
  title={CSympNet-ID: conformal-symplectic map learning for linearly damped Hamiltonian systems},
  author={Gong, Jiale and Jin, Pengzhan and Kuang, Dongyang and Li, Lu and Tang, Yifa},
  journal={arXiv preprint arXiv:2607.03339},
  year={2026}
}

@article{kumar2021rma,
  title={Rma: Rapid motor adaptation for legged robots},
  author={Kumar, Ashish and Fu, Zipeng and Pathak, Deepak and Malik, Jitendra},
  journal={arXiv preprint arXiv:2107.04034},
  year={2021}
}

@article{ye2026world,
  title={World action models are zero-shot policies},
  author={Ye, Seonghyeon and Ge, Yunhao and Zheng, Kaiyuan and Gao, Shenyuan and Yu, Sihyun and Kurian, George and Indupuru, Suneel and Tan, You Liang and Zhu, Chuning and Xiang, Jiannan and others},
  journal={arXiv preprint arXiv:2602.15922},
  year={2026}
}

@book{bardi1997optimal,
  title={Optimal control and viscosity solutions of Hamilton-Jacobi-Bellman equations},
  author={Bardi, Martino and Dolcetta, Italo Capuzzo and others},
  volume={12},
  year={1997},
  publisher={Springer}
}

\end{document}